\documentclass[preprint]{elsarticle}

\usepackage{lineno,hyperref}
\modulolinenumbers[5]

\usepackage{algorithm}
\usepackage{algorithmic}
\usepackage{multirow}
\usepackage{subfig}
\usepackage[english]{babel}
\newtheorem{theorem}{Theorem}
\usepackage{geometry}

\newtheorem{lemma}{Lemma}
\usepackage{amsmath,amssymb,amsfonts}
\usepackage{rotating}
\usepackage{booktabs}
\newcommand\model{HDP }

\journal{Neural Networks}

\begin{document}

\begin{frontmatter}


\title{Heterophilous Distribution Propagation for Graph Neural Networks}
\author[1,2]{Zhuonan Zheng}
\ead{zhengzn@zju.edu.cn}

\author[2,3]{Sheng Zhou \corref{cor1}}
\ead{zhousheng_zju@zju.edu.cn}

\author[1,2]{Hongjia Xu}
\ead{xu_hj@zju.edu.cn}

\author[1,2]{Ming Gu}
\ead{guming444@zju.edu.cn}

\author[1,2]{Yilun Xu}
\ead{yilun.xu@zju.edu.cn}

\author[4]{Ao Li}
\ead{liao.rio@bytedance.com}

\author[5]{Yuhong Li}
\ead{daniel.lyh@alibaba-inc.com}

\author[1,2]{Jingjun Gu}
\ead{gjj@zju.edu.cn}

\author[1,2]{Jiajun Bu}
\ead{bjj@zju.edu.cn}

\address[1]{College of Computer Science, Zhejiang University, Hangzhou, 310027, China}
\address[2]{Zhejiang Provincial Key Laboratory of Service Robot, Zhejiang University, Hangzhou, 310027, China}
\address[3]{School of Software Technology, Zhejiang University, Ningbo, 315048, China}
\address[4]{Bytedance Inc., Hangzhou, 311113, China}
\address[5]{Alibaba Group, Hangzhou, 310052, China}

\cortext[cor1]{Corresponding author}

\begin{abstract}
Graph Neural Networks (GNNs) have achieved remarkable success in various graph mining tasks by aggregating information from neighborhoods for representation learning.
The success relies on the homophily assumption that nearby nodes exhibit similar behaviors, while it may be violated in many real-world graphs.
Recently, heterophilous graph neural networks (HeterGNNs) have attracted increasing attention by modifying the neural message passing schema for heterophilous neighborhoods.
However, they suffer from insufficient \textbf{neighborhood partition} and \textbf{heterophily modeling}, both of which are critical but challenging to break through.
To tackle these challenges, in this paper, we propose heterophilous distribution propagation (HDP) for graph neural networks.
Instead of aggregating information from all neighborhoods, HDP adaptively separates the neighbors into homophilous and heterphilous parts based on the pseudo assignments during training.
The heterophilous neighborhood distribution is learned with orthogonality-oriented constraint via a trusted prototype contrastive learning paradigm. Both the homophilous and heterophilous patterns are propagated with a novel semantic-aware message passing mechanism.
We conduct extensive experiments on 9 benchmark datasets with different levels of homophily. 
Experimental results show that our method outperforms representative baselines on heterophilous datasets.
\end{abstract}


% \begin{highlights}
% \end{highlights}


\end{frontmatter}

% \linenumbers


\section{Introduction}
Graph Neural Networks (GNNs) aim at learning effective representations for graph data, which have demonstrated exceptional performance across a range of graph mining tasks, including node classification \cite{xiao2022graph}, link prediction \cite{lu2011link}, graph classification \cite{cai2018comprehensive}, and anomaly detection \cite{ma2021comprehensive}. The majority of existing GNNs utilize the neural message passing (NMP) schema and aggregate information from the neighborhood for representation learning.
This is predicated on the homophily assumption, which suggests that nodes in close proximity within a graph are likely to exhibit similar behaviors, such as labels and features. 
This assumption has been widely observed in bibliographic graphs \cite{al2022identifying} and online social networks \cite{feng2021botrgcn}. However, many types of graphs challenge the homophily assumption that connected nodes can exhibit \textbf{heterophily} patterns, which presents significant challenges for the application of GNNs.

To tackle this challenge, Heterophily Graph Neural Networks (HeterGNNs) \cite{pei2020geom, zhu2020beyond, bo2021beyond, luan2022revisiting} have attracted increasing attention from both academic and industry communities in the past few years.
Specifically, they have mainly focused on modifying the message passing process by targeting the characteristics of heterophilous graphs from different perspectives.
Among them, early methods primarily adjust the \textbf{scope of message passing}, such as decreasing the proportion of heterophilous neighbors by enlarging the high-order neighborhood \cite{abu2019mixhop}. Other methods alter the \textbf{message passing process itself}, for instance, by assigning varying weights to neighbors based on the similarity of their representations \cite{wang2022powerful}. Further, some methods modify the \textbf{update process}, such as separating the representations of ego nodes and their neighboring nodes \cite{zhu2020beyond}.

\begin{figure}[t]
    \centering
    \subfloat[A toy example of neighborhood partition]
    {\includegraphics[width=.4\columnwidth]{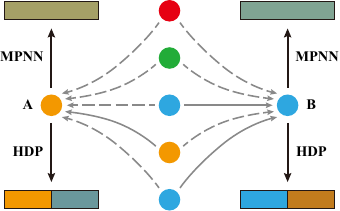}
    \label{fig:intro_a}}
    \hspace{10pt}
    \subfloat[Average connection preference in Cornell.]
    {\includegraphics[width=.55\columnwidth]{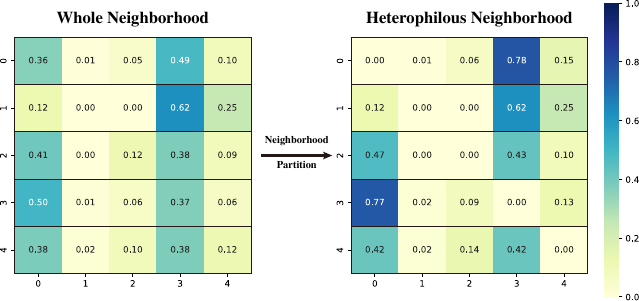}
    \label{fig:intro_b}}
    \caption{The illustration of neighborhood partition and heterophily modeling. (a) A and B are two nodes that belong to different classes but have extremely similar neighborhoods, where colors denote classes. Without neighborhood partition, message passing neural networks(MPNN) produce confused node representations due to the similar neighborhood. On the contrary, partitioning neighborhoods explicitly and handling them separately can increase the discriminability of representations. (b) We count the average connection preference for each class in Cornell. In the whole neighborhood, the connection preferences of different classes are similar, which is probably caused by the imbalanced numbers of nodes in each class. After neighborhood partition, the connection preferences in heterophilous neighborhood shows the discriminability. Thus, heterophily modeling can bring additional information for discriminative representation learning. }
    \label{fig:intro}
\end{figure}

Despite their achievements, we argue that existing methods still suffer from the following shortcomings:
(i) Insufficient \textbf{neighborhood partition}. 
Unlike the homophilous neighborhood that exhibits consistent patterns, the heterophilous neighborhood introduces significantly different patterns and deserves specific modeling. 
However, most existing methods either do not distinguish between homophilous and heterophilous neighborhoods, or simply utilize heuristic thresholds on the representation similarities for naive partition. 
This not only obscures the unique characteristics of heterophilous neighborhoods but also impacts the accurate modeling of homophilous ones, as we show in Figure \ref{fig:intro_a}.
(ii) Insufficient \textbf{heterophily modeling}. As previously mentioned, most existing HeterGNNs have modified the message passing schema and aggregate information from heterophilous neighbors. 
However, given that heterophilous neighbors originate from multiple categories, simply aggregating and fusing heterophily neighbors along with central nodes and homophilous neighbors will lose critical heterophilous connection patterns.
This is a key distinction that separates these methods from those designed for homophily. 
Figure \ref{fig:intro_b} shows the advantages of neighborhood partition and heterophily modeling for real-world datasets.

Although important, overcoming the aforementioned shortcomings is quite challenging:
Firstly, the accurate partition of homophilous and heterphilous neighborhoods relies on the node labels.
Given the limited number of semi-supervised labels, we can only rely on the pseudo labels produced by the HeterGNNs.
The \textbf{mutually dependent} between effective representation learning and neighborhood partition poses significant challenges.
Secondly, heterophily modeling relies on the diverse connections between nodes from different classes.
However, due to the network sparsity and scarcity of node labels, we can not model the heterophily connection patterns for each node from its limited neighborhood.
Instead, the patterns are expected to \textbf{propagate along with the homophilous edges} for more comprehensive modeling.

To tackle the above challenges, in this paper, we propose \textbf{H}eterophilous \textbf{D}istribution \textbf{P}ropagation for Graph Neural Networks (HDP).
More specifically, given a heterophilous graph with an unknown heterophily ratio, HDP first estimates the heterophily level and 
partition the neighborhood according to the semantic assignments, which are dynamically updated along with the training process.
On this basis, we model the heterophilous neighborhood distribution via a simple but effective operator with orthogonality-oriented constraint as a trusted prototype contrastive learning paradigm.
We also theoretically prove the advantages of the operator.
To further enhance the heterophily modeling, we propose a semantic-aware message passing mechanism that propagates both homophilous and heterophilous messages through the edges that connect nodes with the same label.
We conduct extensive experiments on 9 benchmark datasets with different heterophily ratios. Experimental results show that our method outperforms representative baselines on heterophilous datasets.
Our contributions can be summarized as follows:
\begin{itemize}
    \item We point out that existing HeterGNNs suffer from insufficient neighborhood partition and heterophily modeling, which are critical but challenging to break through.
    \item We propose heterophilous distribution propagation for graph neural networks (HDP), which consists of the semantic-aware neighborhood partition and heterophilous neighborhood distribution modeling to address the aforementioned challenges.
    \item We conduct extensive experiments to compare our models against 13 other competitors on 9 benchmark datasets with different levels of homophily.  Experimental results show the superiority of our methods on the heterophilous datasets.
\end{itemize}

\section{Related Work}
\textbf{Graph Neural Networks} have shown great power to model graph structured data. 
The representative designs \cite{kipf2016semi, hamilton2017inductive, velivckovic2017graph, chen2020simple} aim to smooth features across the graph topology or aggregate information from neighbors and then update ego representations, both leading to the similar representations between central nodes and neighbors.
However, most of them are based on an implicit assumption that the graph is homophily, while real-world graphs do not always obey it. This leads to their poor performance on heterophily graphs where nodes of different classes are connected.

\textbf{Heterophilous GNNs}\cite{zheng2022graph, zhu2023heterophily} have been proposed to tackle this problem. 
Some of them tend to \textbf{reduce the negative impact of heterophilous neighbors}.
A naive idea is to decrease the proportion of heterophilous neighbors from the data level, such as neighborhood extension by high-order neighbors and graph reconstruction by node similarities.
MixHop \cite{abu2019mixhop}, a representative method, aggregates messages from multi-hop neighbors to adapt to different scales.
Apart from multi-hop neighbors, UGCN \cite{jin2021universal} and SimP-GCN \cite{jin2021node} extend the neighbor set by adding similar but disconnected nodes through the kNN algorithm.
WRGAT \cite{suresh2021breaking} calculates the structural similarity according to the degree sequence of the neighbors and utilizes it to reconstruct a multi-relational graph.
Geom-GCN \cite{pei2020geom} defines the geometric relationships to discover potential neighbors as a complement to the original neighbor set.
Li et al. \cite{li2023restructuring} learn discriminating node representations with the idea of spectral clustering. Based on the representation distance between nodes, a graph is reconstructed to maximize homophily.
Also, some methods reduce the weights of heterophilous neighbors during aggregation from the model level.
H2GCN \cite{zhu2020beyond} separates ego- and neighbor-representations to prevent ego-node from the pollution of heterophilous neighbors.
The subsequent methods distinguish the neighbors explicitly or implicitly and set the corresponding weights.
GGCN \cite{yan2022two} distinguishes neighbors according to the signs of representation cosine similarity and applies different update weights.
HOG-GCN \cite{wang2022powerful} captures the pair-wise homophily estimation from attribute space and topology space and uses it as the aggregate weight.

Further, some methods found \textbf{the advantages of heterophilous neighbors} and utilized them through high-pass filters \cite{dong2021adagnn, wu2022beyond} or negative aggregation weights \cite{wu2023signed}.
FAGCN \cite{bo2021beyond} learns the attention weights of low- and high-frequency signals for each node, corresponding to the negative-available aggregate weights in the spatial domain.
On this basis, ACM-GCN \cite{luan2022revisiting} introduces the identity filter to capture more information about the original feature.
GBK-GNN \cite{du2022gbk} utilizes two kernels to capture homophilous and heterophilous information and selects the result by a gate for each node.
Similarly, GloGNN \cite{li2022finding} learns a coefficient matrix based on the self-expressiveness assumption of the linear subspace model, which guides the global message passing.

\section{Preliminaries}
In this section, we first give the notations and problem description, then introduce the concepts used in this paper.

\subsection{Notations}
Let $\mathcal{G=(V,E)}$ be an undirected graph with nodes $\mathcal{V}$ and edges $\mathcal{E}$. $N$ denotes the number of nodes.
$\mathbf{A} \in \mathbb{R}^{N\times N}$ is the adjacency matrix and $\mathbf{X} \in \mathbb{R}^{N\times F}$ is the node feature matrix with feature dimension $F$. 
Node labels are represented as $\mathbf{Y} \in \mathbb{R}^{N \times 1}$, and only a part of it is available.
% $\mathcal{N}$ denotes the neighborhoods, where $\mathcal{N}_i$ means the neighbor set of node $i$.  
We use $K$ as the number of classes and $D$ as the representation dimension.
$\mathbf{S}^{tra}$, $\mathbf{S}^{tar}$ and $\mathbf{S}^{tru}$ denote the training set, target set for structure encoding and trust set for trusted prototype contrastive loss respectively, which are described in the following section.

\subsection{Problem Description}
In this paper, we mainly focus on graph representation learning, of which the performance is evaluated by the semi-supervised node classification task. Specifically, in a graph $\mathcal{G=(V,E)}$, each node belongs to one of $K$ classes and a part of labels $\mathbf{Y}$ are already known. The objective is to predict the labels of other nodes.

\subsection{Homophily and Heterophily}
Homophily and Heterophily are two opposite concepts related to edges, features and labels of a graph.
We use the edge homophily ratio $h=\frac{|{(u,v)|(u,v)\in \mathcal{E}} \ \wedge \  y_u=y_v|}{|\mathcal{E}|} \in [0,1]$, the proportion of edges connecting nodes of the same class, to measure the specific homophily level for a graph. 
Graphs with strong homophily tend to have a high $h$ close to 1, while graphs with strong heterophily are the opposite, i.e. $h \rightarrow 0$.

\section{Methodology}

\begin{figure*}[t]
\centerline{\includegraphics[width=\textwidth]{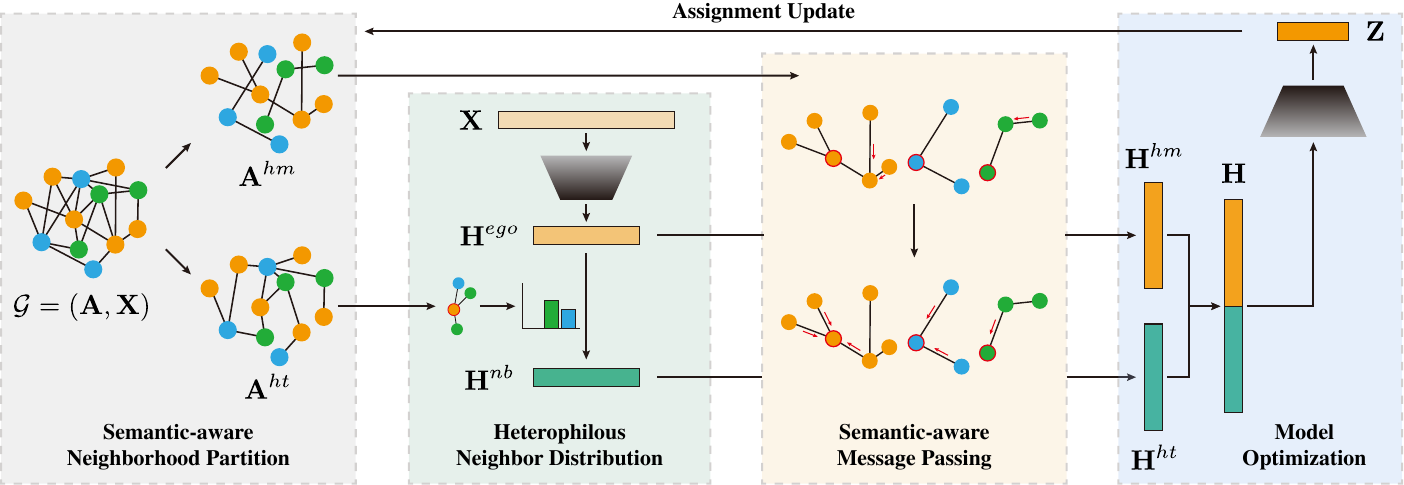}}
\caption{Overall framework of HDP, which contains three main parts including semantic-aware neighborhood partition, heterophilous neighbor distribution modeling and semantic-aware message passing.
}
\label{framework}
\end{figure*}

In this section, we introduce our proposed \model model in detail. 
An overview of \model is given in the Figure \ref{framework}. 
\model first estimates the heterophily level of a graph and partitions the neighborhood as homophilous and heterophilous ones according to the semantic assignments, which are constantly refined during the training process. 
Then, the heterophilous neighbor distribution is modeled for each node based on the heterophilous neighborhood and an orthogonality-oriented constraint.
Further, \model propagates messages via a semantic-aware message passing mechanism to capture class-level information approximatively.

\subsection{Semantic-Aware Neighborhood Partition}
Most of the existing methods distinguished the neighborhood by the representation similarity and a manually selected threshold \cite{yan2022two, liu2023beyond, choi2022finding}. 
However, this approach lacks explainability since the definition of homophily is based on semantic labels instead of representation. 
Hence, we propose a semantic-aware neighborhood partition mechanism based on the soft assignments, which is a special form of labels.
Specifically, we first estimate the level of heterophily in the graph as the guidance and then partition the neighborhood.

\subsubsection{Heterophily Estimation}
We use the homophily ratio to measure how heterophily the graph is that a lower homophily ratio means stronger heterophily. 
To estimate the homophily ratio of a graph, we start with an assumption that the training set shares a similar edge distribution with the full graph, which means the homophily ratio won't be too far from the truth if there are enough nodes in the training set. 
Therefore, the homophily ratio of the full graph can be estimated from the training set where a part of labels are available:
\begin{equation}
    h' = \frac{| \{(u,v)| (u,v) \in \mathcal{E}'  \ \wedge \ y_u = y_v \ \wedge \ u,v \in \mathbf{S}^{tra}\} |}{| \{(u,v)| (u,v) \in \mathcal{E}' \ \wedge \ u,v \in \mathbf{S}^{tra} \} |} ,
    \label{eq:homo1}
\end{equation}
where $\mathcal{E}' \in \{\mathcal{E}, \mathcal{E}^2\}$ is the partitioned neighborhood, which can be 1-hop or 2-hop considering efficiency and the number of isolated nodes in results, $\mathbf{S}^{tra}$ denotes the training set.
As the estimated homophily ratio $h'$ can't be completely accurate, we slightly rescale it to find a suitable boundary for neighborhood partition:
% find a suitable boundary for partition:
\begin{equation}
    \widehat{h} = \lambda h',
    \label{eq:homo2}
\end{equation}
where $\lambda\in [0.8, 1.2]$ is a parameter that controls the direction and strength of rescaling.

\subsubsection{Neighborhood Partition}
Soft assignment $\mathbf{Z} \in \mathbb{R}^{N \times K}$ is the result predicted by \model, where the sum of each row is 1 and element $\mathbf{z}_{ij}$ in row $i$ and column $j$ indicates the probability of node $i$ belonging to class $j$.
On this basis, we can directly calculate the probabilities that two nodes belong to the same class, which are also equivalent to the homophilous edge probabilities $\mathbf{P}$:
\begin{equation}
    \mathbf{P}_{uv}=
    \begin{cases}
         \mathbf{z}_u \mathbf{z}_v^T, & (u,v) \in \mathcal{E}',  \\
         0, & \text{otherwise}.
    \end{cases}
\end{equation}
The neighborhood is then partitioned to fit the estimated heterophily level according to $\mathbf{P}$, i.e. a lower homophily ratio corresponds to fewer homophilous edges and vice versa.
Specifically, \model automatically generate a threshold $\epsilon$ based on $\widehat{h}$:
\begin{equation}
    \epsilon = \text{TopK}(\mathbf{P}, \widehat{h}|\mathcal{E}'|) ,
\end{equation}
where $\text{TopK}(\mathbf{x},y)$ means the $y$-largest element in $\mathbf{x}$. 
Then the homophilous neighborhood $\mathbf{A}^{hm}$ and heterophilous neighborhood $\mathbf{A}^{ht}$ can be partitioned by threshold filtering:
\begin{equation}
    \mathbf{A}^{hm}_{uv} = 
    \begin{cases}
         1, & \mathbf{P}_{uv}\geq \epsilon\ \wedge\ (u,v) \in \mathcal{E}',   \\
         0, & \text{otherwise}.
    \end{cases} 
    \label{eq:parthm}
\end{equation}
\begin{equation}
    \mathbf{A}^{ht}_{uv} = 
    \begin{cases}
         1, & \mathbf{P}_{uv} < \epsilon\ \wedge\ (u,v) \in \mathcal{E}',  \\
         0, & \text{otherwise}.
    \end{cases} 
    \label{eq:partht}
\end{equation}

To avoid the defects of outdated partition results, we conduct an \textbf{update strategy} that refines the partition results dynamically when the assignments become more accurate during the training process.

\subsection{Heterophilous Neighborhood Modeling}
The heterophilous neighborhood deserves to be modeled separately rather than integrated to the central nodes, since its diverse preference is crucial for the discriminability of representations.
Thus, we model the heterophilous neighborhood in three steps: 
(1) constructing ego representation for each node,
(2) modeling the heterophilous neighbor distribution of each node,
and (3) propagating them via a semantic-aware message passing mechanism.

\subsubsection{Ego Representation Construction}
Firstly, we construct a representation for each node which contains the attribute and structural information of node ego.

\begin{figure}
    \centering
    \includegraphics[width=0.9\textwidth]{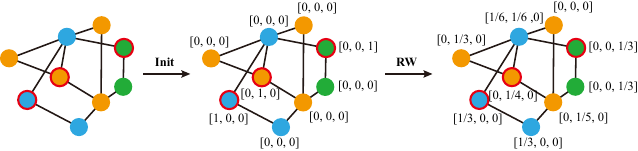}
    \caption{The illustration of semantic structural encoding. Nodes with a red circle denote the target nodes. After initialization, the structural embedding is calculated through the random walk with self-loop.}
    \label{fig:sse}
\end{figure}

To better utilize the topological information without considering the distinguishing of edges, we introduce \textbf{Semantic Structural Encoding (SSE)} as a supplementary feature alongside node attributes.
Distance Encoding \cite{li2020distance} is used where the embedding $\mathbf{X}^{str}$ denotes the landing probabilities of random walks from the corresponding node to some target nodes. 
Different from the existing methods that select target nodes mainly according to topology or randomness \cite{lu2021graph, you2019position}, we construct a target node set $\mathbf{S}^{tar}$ with the nodes in the training set $\mathbf{S}^{tra}$ to implicitly introduce some semantic information: $\mathbf{S}^{tar}\subseteq  \mathbf{S}^{tra}$.
First, the structural embedding $\hat{\mathbf{X}}^{str}_i$ is initialized as a unique one-hot vector for nodes in the target set and an all-zero vector otherwise.
\begin{equation}
    \hat{\mathbf{X}}^{str}_i=
    \left\{
    \begin{array}{cc}
         \left[ 0,0,...,1,...,0 \right], & i \in \mathbf{S}^{tar}  \\
         \left[ 0,0,...,0,...,0 \right], & i \not\in \mathbf{S}^{tar}
    \end{array} 
    \label{eq:xstr1}
    \right . ,
\end{equation}
where the position of ‘1’ in the one-hot vector is equal to the ranking of $i$ in $S^{tar}$.
The final structural embedding $\mathbf{X}^{str}$ can be calculated by the transfer probability of random walks:
\begin{equation}
    \mathbf{X}^{str} = (\mathbf{D}^{-1}\mathbf{A})^{\kappa}\hat{\mathbf{X}}^{str},
    \label{eq:xstr2}
\end{equation}
where $\kappa$ is the hop number of random walk. 
Finally, the structural embedding captures the topological information about target nodes of different classes around the central node, which also brings semantic information. 
Figure \ref{fig:sse} gives a toy example of semantic structural encoding.

The ego representations of nodes are obtained by a simple MLP with nodes' features and structural embeddings as input:
\begin{equation}
    \mathbf{H}^{ego} = \text{MLP}([\mathbf{X}\|\mathbf{X}^{str}]),
    \label{eq:ego}
\end{equation}
where $[\cdot \| \cdot]$ denotes the concatenation operation.

\subsubsection{Heterophilous Neighbor Distribution}
To model the distribution of heterophilous neighbors, it would be sufficient to design an operator (i.e. function) that takes heterophilous neighbors as input, with each distinct distribution of heterophilous neighbors corresponding to a different output $o$ in the output space $O$, i.e.
\begin{equation}
\begin{aligned}
    O=\{o_i &\mid o_i=\text{Operator}(\{\mathbf{h}^{ego}_j|{\mathbf{A}^{ht}_{ij}=1}\}), i \in \mathbb{N}\}\\
    &\text{s.t. }\forall \ o_i, o_j \in O, \ o_i \neq o_j \text{ w.r.t. } i \neq j 
\end{aligned}.
\end{equation}

On this basis, we find that the distribution of heterophilous neighbors can be effectively modeled by the mean operator, with the following two assumptions:
(1) Assuming node representations exhibit clustering characteristics, where the average distance within a class is significantly smaller than the average distance between different classes.
This implies that the representations of similar nodes are linearly correlated within a certain range of error.
(2) Assuming the existence of a clustering center for each class's representations, referred to as a prototypes $\{\mathbf{c}_k| k \in K\}$.
We have the following theorem with detailed proof:
{
\linespread{1.2} 
\begin{theorem}
\label{the:1}
Let $\text{Mean}(\{\mathbf{h}^{ego}_j|{\mathbf{A}^{ht}_{ij}=1}\})$
be mean operator that aggregate heterophilous neighbor representations, $\mathbf{c}_{k}$ be the prototype of $k$-th class. Function $\text{Mean}(\{\mathbf{h}^{ego}_j|{\mathbf{A}^{ht}_{ij}=1}\})$ is injective if it is satisfied that all class prototypes $\mathbf{c}_{k}$ are orthogonal to each other.  
\end{theorem}
}
The injectivity ensures that each element in the domain of the input (i.e. heterophilous neighbors' distribution) has a distinct and unique output in the output domain. We found that as long as the conditions of Theorem \ref{the:1} are satisfied, the mean operator can be regarded as an injective function within a certain range of error, a simple yet effective approach to perform heterophilous neighborhood modeling.

To prove that heterophily neighbor patterns can be modeled by the mean operator, it suffices to demonstrate that the mapping function from the mean of $\{\mathbf{h}^{ego}_j|{\mathbf{A}^{ht}_{ij}=1}\}$ to $H^{nb}$ in terms of embedding is injective.

\begin{lemma}
    Injectivity is equivalent to null space equals $\{0\}$.
Let $T \in \mathcal{L}(V, W)$, $T(v)=T \cdot v=w$. Then $T$ is injective if and only if $\text{null} (T)=\{0\}$.
\end{lemma}
\textbf{Proof of lemma 1:}
\textbf{Sufficiency}: First, suppose $T$ is injective. We want to prove that null $T=\{0\}$. We already know that $\{0\} \subset$ $\text{null}(T)$. To prove the inclusion in the other direction, suppose $v \in \text{null}(T)$, then
$
T(v)=0=T(0) .
$
Because $T$ is injective, the equation above implies that $v=0$. Thus we can conclude that null $T=\{0\}$, as desired. \textbf{Necessity}: To prove the implication in another direction, now suppose $\text{null}(T)=\{0\}$. We want to prove that $T$ is injective. To do this, suppose $u, v \in V$ and $T (u)=T (v)$. Then
\begin{equation}
    0=T(u)-T(v)=T(u-v) .    
\end{equation}
Thus $u-v$ is in null $T$, which equals $\{0\}$. Hence $u-v=0$, which implies that $u=v$. Hence $T$ is injective, as desired.

Having the \textbf{Lemma 1} proofed, now we express the mean operator in the following form: 
\begin{equation}
    \textbf{M}\mathbf{X}=b,
\end{equation}
where $M_{1*n}$ represents the mean operator, $\mathbf{X}_{n*D}$ is the matrix formed by embeddings of heterophilous neighbors, and $b$ is the resulting new embedding. Assuming that embeddings of the same type of heterophilous neighbors are linearly dependent, we can rewrite this equation as:
\begin{equation}
    \textbf{M}^{'}\mathbf{X}_p\approx b ,
\end{equation}
where $\textbf{M}_{1*K}^{'}$ is a weighted mean operator, $\mathbf{X}_p$ is a $K*D$ prototype embedding matrix, $K$ is the number of classes. The injectivity of mean operator $\textbf{M}$ involves considering the solution for $\textbf{M}^{'}\mathbf{X}_p = 0$. 
It is clear that if it is satisfied that all $\mathbf{X}_{p}^{k}$ are orthogonal to each other, the null space of $\textbf{M}^{'} = \{0\}$, indicating that the mean operator is approximately injective. In other words, each distinct input produces a unique output. Thus in this manner, the mean operator applied to heterogeneous neighbors can generate distinguishable embeddings based on the distribution of heterogeneous neighbors.

Hence, we approximatively model the heterophilous neighbor distribution by the mean operator with a \textbf{orthogonality-oriented constraint} to make the prototypes as orthogonal as possible, which is described in Sec \ref{sec:opt}.
The heterophilous neighbor distributions are formatted as follows:
\begin{equation}
    \mathbf{H}^{nb} = {\mathbf{D}^{ht}}^{-1} \mathbf{A}^{ht} \mathbf{H}^{ego} .
    \label{eq:nb}
\end{equation}
where $\mathbf{D}^{ht}$ is degree matrix with entries $\mathbf{D}^{ht}_{ii}=\sum_j \mathbf{A}^{ht}_{ij}$.

\subsubsection{Semantic-Aware Message Passing}
Due to the sparse nature of graph data, the heterophilous neighbor distribution of a single node could be chaotic. 
Hence, we introduce \textbf{Semantic-aware Message Passing (SMP)} based on the homophilous neighborhood, which aggregates information from different hops of homophilous neighbors with adaptive weights.
Intuitively, nodes with similar semantics share similar characters of not only nodes themselves but also neighbor distributions.
Thus, aggregating information from enough homophilous neighbors can approximatively model the class-level heterophilous neighborhood, which is more accurate and discriminative.
Specifically, SMP is a multilayer module in which the messages are propagated only on the homophilous neighborhood.
The $l$-th layer of SMP is expressed as follows:
\begin{equation}
\begin{split}
    \widetilde{\mathbf{H}}^{l} &= \mathbf{D}^{hm^{-1}}\mathbf{A}^{hm} \mathbf{H}^{(l-1)},\\
    \alpha^{l} &= f_{\varphi^l}([\mathbf{H}^{0} \| \widetilde{\mathbf{H}}^{l}]) , \\
    \mathbf{H}^l &= \alpha^{l} \mathbf{H}^0 + (1-\alpha^{l)}) \widetilde{\mathbf{H}}^{l},
\end{split}
\end{equation}
where $\mathbf{D}^{hm}$ is a degree matrix with entries $\mathbf{D}^{hm}_{ii}=\sum_j \mathbf{A}^{hm}_{ij}$, $\mathbf{H}^{0}$ is the input of whole SMP, and the message from neighbors and ego nodes are linearly combined in which the weights $\alpha^l\in \mathbb{R}^{N \times 1}$ are set by a weight learner $f_{\varphi^l}$.

Then, we propagate the heterophilous neighbor distribution via SMP to capture the heterophilous distribution representations:
\begin{equation}
    \mathbf{H}^{ht} = \text{SMP}(\mathbf{H}^{nb}, l^{ht}),
    \label{eq:smpht}
\end{equation}
where $l^{ht}$ is the number of SMP layers. 
Meanwhile, propagating the nodes' ego representations via SMP can capture the homophilous representations:
\begin{equation}
    \mathbf{H}^{hm} = \text{SMP}(\mathbf{H}^{ego}, l^{hm}).
    \label{eq:smphm}
\end{equation}

Now we have two kinds of representations that capture homophilous and heterophilous neighborhood information respectively.
The final node representations are the concatenation of $\mathbf{H}^{hm}$ and $\mathbf{H}^{ht}$, which can be the input to other downstream tasks.
\begin{equation}
    \mathbf{H} = [\mathbf{H}^{hm} \| \mathbf{H}^{ht}] .
    \label{eq:geth}
\end{equation}
For node classification, the soft assignments $\mathbf{Z}$ and predicted labels $\widehat{\mathbf{Y}}$ are given by a classifier $f_{\psi}$:
\begin{equation}
    \mathbf{Z} = f_{\psi}(\mathbf{H}),\ \widehat{\mathbf{Y}} = \arg\max(\mathbf{Z}) .
    \label{eq:getz}
\end{equation}

\subsection{Model Training}
\label{sec:opt}
We introduce two modules for model training, including assignment initialization and optimization. 
The former provides the assignment for the first neighborhood partition while the latter describes the optimization object of the whole model.

\subsubsection{Assignment Initialization}
\model partitions the neighborhood by the semantic assignments, which need to be initialized to obtain a relatively accurate result at the beginning of training.
For the assignment initialization, a naive approach is to train a classifier with only nodes' ego features. 
However, the neighbor features may also provide helpful information for classification.
Since we don't know the homophily ratio of the graph, we separate the node's ego features $\mathbf{X}$ and the corresponding neighbor features $\mathbf{X}^{nb}$ to avoid pollution:
\begin{equation}
    % \mathbf{X}^{nb} = \frac{\sum_{j\in\mathcal{N}_{i}} X_j}{|\mathcal{N}_{i}|} .
    \mathbf{X}^{nb} = \widehat{\mathbf{A}} \mathbf{X},
    \label{eq:xnb}
\end{equation}
where $\widehat{\mathbf{A}}$ is the normalized adjacency matrix.
The new node features are obtained through concatenating the ego features, neighbor features and structural embedding:
\begin{equation}
    \mathbf{X}^{all} = [\mathbf{X} \| \mathbf{X}^{nb} \| \mathbf{X}^{str}] .
\end{equation}
In practice, we choose some of them to construct the new node features according to the performance on the validation set.
Finally, we can get the soft assignments $\mathbf{Z}$ through an MLP classifier trained by cross-entropy loss:
\begin{equation}
    \mathbf{Z} = \text{MLP}^{init}(\mathbf{X}^{all}) .
    \label{eq:init}
\end{equation}
To avoid errors caused by precision, we rescale the assignments to a relatively high level during the neighborhood partition.

\subsubsection{Optimization}
\model contains two kinds of objectives: a commonly used cross-entropy loss for node classification and a Trusted Prototype Contrastive (TPC) loss as the orthogonality-oriented constraint for class prototypes.

The cross-entropy function can measure the gap between predicted results and the ground truth:
\begin{equation}
    \mathcal{L}^{ce} = \text{CE}(\mathbf{Z}, \mathbf{Y}) ,
    \label{eq:celoss}
\end{equation}
where $\text{CE}(\cdot,\cdot)$ denotes the cross-entropy function.

In order to constrain the orthogonality of class prototypes, we introduce the Trusted Prototype Contrastive (TPC) loss which is inspired by the original prototype contrastive learning (PCL) \cite{li2020prototypical}.
Contrastive learning \cite{chen2020simple} aims to pull positive samples together while pushing negative samples away. 
As a variant, PCL constructs positive and negative samples between samples and class prototypes calculated by the pseudo labels, leading to highly discriminative representations. 
In our TPC loss, we first select some high confidence nodes as trust set $\mathbf{S}^{tru}$:
\begin{equation}
    \mathbf{S}^{tru} = \{v_i| \mathbf{Z}^{max}_i \geq \delta\} ,
    \label{eq:trust1}
\end{equation}
where $\mathbf{Z}^{max}$ is the maximum value in each row of $\mathbf{Z}$, denoting the largest probability of each node belonging to any class. $\delta$ is a threshold decided by the accuracy $\rho$ of training and validation set:
\begin{equation}
    \delta = \text{TopK}(\mathbf{Z}^{max}, \rho|\mathcal{V}|) .
\end{equation}
Then, the prototype of class $j$ can be calculated as the mean of node ego representations $\mathbf{H}^{ego}$ within trust set:
\begin{equation}
\begin{split}
    \mathbf{c}_j &= \frac{1}{|\mathbf{S}^{tru}_j|}\sum_{v_i \in \mathbf{S}^{tru}_j} \mathbf{h}^{ego}_{i}, \\ 
    \mathbf{S}^{tru}_j &= \{v_i|v_i \in \mathbf{S}^{tru} \wedge \widehat{\mathbf{Y}}_i = j\}.
\end{split}
\end{equation}
Formally, the trusted prototype contrastive loss can be expressed as follows:
\begin{equation}
    \mathcal{L}^{tpc} = -\sum_{v_i \in \mathbf{S}^{tru}} \log\frac{\exp(s(\mathbf{h}^{ego}_{i},\mathbf{c}_j)/\tau)}{\sum_{k=1}^K\exp([s(\mathbf{h}^{ego}_{i},\mathbf{c}_k)]_{+}/\tau))} , 
    \label{eq:tscloss}
\end{equation}
where $\tau$ is a temperature parameter, $[\cdot]_{+}=\max(\cdot, 0)$ and $s(\cdot,\cdot)$ is cosine similarity function:
\begin{equation}
    s(\mathbf{h}^{ego}_{i}, \mathbf{c}_j) = \frac{\mathbf{h}^{ego}_{i} \cdot \mathbf{c}_j}{|\mathbf{h}^{ego}_{i}||\mathbf{c}_j|} .
\end{equation}
In the ideal case, the cosine similarity will be 1 between representations and positive prototypes and 0 between representations and negative prototypes since it's set to be non-negative during optimization.
In other words, the optimal case of TPC loss is that all class prototypes are orthogonal to each other since the prototypes are made of representations, indicating that the TPC loss is an orthogonality-oriented constraint as we desired.
Meanwhile, it also pulls nodes from the same class together and pushes nodes from different classes away, which guarantees the first assumption of heterophilous distribution modeling and brings discriminability to the representations.

The overall optimization objective can be written as follows:
\begin{equation}
    \mathcal{L} = \mathcal{L}^{ce} + \beta\mathcal{L}^{tpc} .
    \label{eq:allloss}
\end{equation}
where $\beta$ is a weight parameter.
Finally, we summarize the whole process of \model in Algorithm \ref{alg}.

\begin{algorithm}
\caption{Algorithm of HDP}
\label{alg}
\begin{algorithmic}[1]
    \REQUIRE Graph $\mathcal{G=(V,E)}$, training set $\mathbf{S}^{tra}$, node labels $\mathbf{Y}$, adjacency matrix $\mathbf{A}$, node features $\mathbf{X}$, rescaling parameter $\lambda$, epoch $E$
    \ENSURE Predicted labels $\hat{\mathbf{Y}}$ 
    \STATE Construct structural embedding $\mathbf{X}^{str}$ via Eq.\ref{eq:xstr1} and Eq.\ref{eq:xstr2}.
    \STATE Initialize the assignment $\mathbf{Z}$ via Eq.\ref{eq:init}.
    \STATE Estimate the homophily ratio of graph via Eq.\ref{eq:homo1} and Eq.\ref{eq:homo2}.
    \STATE Partition the neighborhood to $\mathbf{A}^{hm}$ and $\mathbf{A}^{ht}$ via Eq.\ref{eq:parthm} and Eq.\ref{eq:partht}.
    \STATE Establish trust set $S^{tru}$ via Eq.\ref{eq:trust1}.
    \FOR{iteration 1, 2, ..., E}
        \STATE Construct ego representations $\mathbf{H}^{ego}$ for nodes via Eq.\ref{eq:ego}.
        \STATE Modeling heterophilous neighbor distribution $\mathbf{H}^{nb}$ via Eq.\ref{eq:nb}.
        \STATE Propagate $\mathbf{H}^{nb}$ to capture the heterophilous distribution representations $\mathbf{H}^{ht}$ via Eq.\ref{eq:smpht}
        \STATE Propagate $\mathbf{H}^{ego}$ to capture the homophilous representations $\mathbf{H}^{hm}$ via Eq.\ref{eq:smphm}
        \STATE Obtain final representations $\mathbf{H}$, assignments $\mathbf{Z}$ and predicted labels $\hat{\mathbf{Y}}$ via Eq.\ref{eq:geth} and Eq.\ref{eq:getz}.
        \STATE Calculate loss $\mathcal{L}$ via Eq.\ref{eq:tscloss} and Eq.\ref{eq:allloss}.
        \STATE Back-propagation $\mathcal{L}$ to optimize the weights of networks.
        \IF{current assignment $\mathbf{Z}$ performs better}
            \STATE update the neighborhood partition results $\mathbf{A}^{hm}$ and $\mathbf{A}^{ht}$ via Eq.\ref{eq:parthm} and Eq.\ref{eq:partht} with current $\mathbf{Z}$.
            \STATE update the trust set $\mathbf{S}^{tru}$ via Eq.\ref{eq:trust1} with current $\mathbf{Z}$.
        \ENDIF
    \ENDFOR
    \RETURN $\hat{\mathbf{Y}}$
\end{algorithmic}
\end{algorithm}

\section{Experiments}
In this section, we first evaluate the representation learning performance of \model through node classification task against some state-of-the-art methods on 9 public datasets. 
Then, the effectiveness of components in \model is shown by an ablation study and some visualizations.

\begin{sidewaystable}[!htbp]
\scriptsize
% \footnotesize
\centering
\caption{Detailed statistics of datasets and classification performance on 6 datasets with various levels of heterophily and 3 homophilous datasets.}
\renewcommand{\arraystretch}{1.2}
\setlength{\tabcolsep}{3pt}
\begin{tabular}{c|cccccc|ccc} 
\toprule
Methods & \textbf{Cornell} & \textbf{Texas} & \textbf{Wisconsin} & \textbf{Chameleon} & \textbf{Actor} & \textbf{Squirrel} & \textbf{Cora} & \textbf{Citeseer} & \textbf{Pubmed}  \\
Homo. Ratio & 0.3 & 0.11 & 0.21 & 0.23 & 0.22 & 0.22 & 0.81 & 0.74 & 0.8 \\
\#Nodes & 183 & 183 & 251 & 2277 & 7600 & 5201 & 2708 & 3327 & 19717 \\
\#Edges & 280 & 295 & 466 & 31421 & 26752 & 198493 & 5278 & 3703 & 44327 \\
\#Features & 1703 & 1703 & 1703 & 2325 & 931 & 2089 & 1433 & 3703 & 500 \\
\#Classes & 5 & 5 & 5 & 5 & 5 & 5 & 7 & 6 & 3 \\
\midrule
 MLP & 84.32 ± 4.32 & 82.43 ± 6.76 & 85.10 ± 3.53 & 48.27 ± 1.91 & 36.70 ± 1.09 & 34.30 ± 0.99 & 73.76 ± 1.95 & 72.87 ± 2.07 & 87.39 ± 0.34 \\ 
\midrule
GCN & 61.08 ± 5.43 & 62.16 ± 4.83 & 60.98 ± 5.15 & 67.19 ± 2.16 & 30.31 ± 1.19 & 51.71 ± 1.18 & 86.94 ± 0.97 & 75.92 ± 1.49 & 87.64 ± 0.54 \\
GAT & 59.46 ± 3.63 & 64.59 ± 5.47 & 60.78 ± 8.68 & 63.38 ± 1.25 & 30.37 ± 0.86 & 54.07 ± 1.85 & 85.37 ± 1.46 & 74.92 ± 1.70 & 86.38 ± 0.48 \\
GCNII & 77.86 ± 3.79 & 77.57 ± 3.83 & 80.39 ± 3.40 & 63.86 ± 3.04 & 37.44 ± 1.30 & 38.47 ± 1.58 & \textbf{88.37 ± 1.25} & \underline{77.33} ± 1.48 & \textbf{90.15 ± 0.43} \\
\midrule
MixHop & 81.08 ± 6.28 & 82.97 ± 6.17 & 84.31 ± 3.16 & 66.18 ± 1.57 & 34.81 ± 0.58 & 56.26 ± 1.69 & 86.48 ± 1.04 & 76.72 ± 1.07 & 88.39 ± 0.45 \\
H$_2$GCN & 82.16 ± 4.80 & 84.86 ± 6.77 & 86.67 ± 4.69 & 59.39 ± 1.98 & 35.86 ± 1.03 & 37.90 ± 2.02 & 87.67 ± 1.42 & 77.07 ± 1.64 & 89.59 ± 0.33 \\ 
UGCN & 73.78 ± 5.41 & 75.95 ± 5.85 & 78.43 ± 4.96 & 63.22 ± 2.01 & 32.55 ± 1.36 & 50.09 ± 2.91 & 72.43 ± 2.75 & 74.53 ± 1.38 & 80.73 ± 1.36 \\ 
WRGAT & 81.62 ± 3.90 & 83.62 ± 5.50 & 86.98 ± 3.78 & 65.24 ± 0.87 & 36.53 ± 0.77 & 48.85 ± 0.78 & 88.20 ± 2.26 & 76.81 ± 1.89 & 88.52 ± 0.92 \\
GPR-GNN & 79.46 ± 6.30 & 83.51 ± 5.98 & 83.53 ± 4.04 & 69.08 ± 2.41 & 35.22 ± 1.00 & 52.06 ± 1.31 & 87.77 ± 1.14 & 75.98 ± 1.53 & 86.98 ± 0.49 \\ 
LINKX & 77.84 ± 5.81 & 74.60 ± 8.37 & 75.49 ± 5.72 & 68.42 ± 1.38 & 36.10 ± 1.55 & \underline{61.81 ± 1.80} & 84.64 ± 1.13 & 73.19 ± 0.99 & 87.86 ± 0.77 \\
GGCN & 85.68 ± 6.63 & 84.86 ± 4.55 & 86.86 ± 3.29 & 71.14 ± 1.84 & \underline{37.54 ± 1.56} & 55.17 ± 1.58 & 87.95 ± 1.05 & 77.14 ± 1.45 & 89.15 ± 0.37 \\
ACM-GCN & 85.14 ± 6.07 & \underline{87.84 ± 4.40} & \underline{88.43 ± 3.22} & 69.14 ± 1.91 & 36.63 ± 0.84 & 55.19 ± 1.49 & 87.91 ± 0.95 & 77.32 ± 1.70 & \underline{90.00 ± 0.52} \\
GloGNN & \underline{85.95 ± 5.10}  & 84.32 ± 4.15 & 88.04 ± 3.22 & \underline{71.21 ± 1.84} & \textbf{37.70 ± 1.40} & 57.88 ± 1.76 & \underline{88.33 ± 1.09} & \textbf{77.41 ± 1.65} & 89.62 ± 0.35 \\
\midrule
% \model & \textbf{87.57 ± 2.16} & \textbf{88.38 ± 3.43} & \underline{88.43 ± 3.77} & \textbf{72.21 ± 2.22} & 37.14 ± 1.26 & \underline{61.73 ± 1.03} & 87.00 ± 1.65 & 77.11 ± 1.39 & 89.59 ± 0.38 \\
\model & \textbf{87.84 ± 4.23} & \textbf{88.38 ± 4.20} & \textbf{88.82 ± 3.40} & \textbf{71.56 ± 2.52} & 37.26 ± 0.67 & \textbf{62.07 ± 1.57} & 87.00 ± 1.35 & 77.10 ± 1.56 & 89.49 ± 0.47 \\
\bottomrule
\end{tabular}
\label{tbl:per}
\end{sidewaystable}

\subsection{Datasets and Baselines}
Experiments are conducted on 6 public heterophilous graph datasets including Cornell, Texas, Wisconsin, Chameleon, Actor, and Squirrel~\cite{pei2020geom}, and 3 homophilous datasets including Cora, Citeseer and Pubmed~\cite{yang2016revisiting}.
The detailed statistics of these datasets are summarized in Table \ref{tbl:per} while the descriptions are in follows:
\begin{itemize}
    \item Cornell, Texas and Wisconsin are three sub-datasets of the webpage dataset collected from computer science departments of various universities, where nodes are web pages belonging to one of five categories, and edges represent the hyperlinks between them.
    \item Chameleon and Squirrel are two webpage networks in Wikipedia. The nodes are classified into five categories based on their average amounts of monthly traffic.
    \item Actor (also named Film) is a subgraph of the film-director-actor-writer network, where nodes are actors and edges denote the co-occurrence relation between them in Wikipedia pages. 
    \item Cora, Citeseer and Pubmed are citation networks with high homophily. In these datasets, nodes represent the scientific papers while edges denote citations. The node label is the research field of a paper.
\end{itemize}

We compare \model with 13 baseline methods, including (1) MLP; (2) general GNN methods: GCN \cite{kipf2016semi}, GAT \cite{velivckovic2017graph} and GCNII \cite{chen2020simple}; (3) methods adapted to heterophilous graphs: MixHop \cite{abu2019mixhop}, H$_2$GCN \cite{zhu2020beyond}, UGCN \cite{jin2021universal}, WRGAT \cite{suresh2021breaking}, GPR-GNN \cite{chien2020adaptive}, LINKX \cite{lim2021large}, GGCN \cite{yan2022two}, ACM-GCN \cite{luan2022revisiting} and GloGNN \cite{li2022finding}. 
The first six heterophilous GNN methods tend to reduce the negative impact of heterophilous neighbors while the last three utilize the difference between ego node and heterophilous neighbors.

\subsection{Experimental Settings}
We implement \model by PyTorch and run experiments on the Nvidia RTX 3090 GPU. 
The models are optimized by Adam \cite{kingma2014adam}. 
For the hyperparameter setting, we use an anneal strategy to turn the hyperparameter combination based on the results of the validation set. 
Early stop strategy is applied for model training with the parameter "patience".
The assignment initialization is separated from the main part of \model for parameter tuning.
Detailed search space of hyperparameters is listed in Table \ref{tbl:hyper}, while specific settings can be seen in codes.

For a fair comparison, we run the experiments with the same split (48\%/32\%/20\% of nodes for train/validation/test) of datasets from previous papers \cite{pei2020geom,li2022finding}, and report the average accuracy and corresponding standard deviation score over 10 runs on different splits. 
Since the results of some baseline methods on these datasets are public, we directly report them.
For methods with absent results on some datasets, we use the official code released by corresponding authors and finetune the parameters as suggested in the original paper.

\begin{table}[t]
\caption{The search space and settings of hyper-parameters.}
\label{tbl:hyper}
\footnotesize
\renewcommand{\arraystretch}{0.8}
\begin{center}
\begin{tabular}{c|c}
\toprule
\textbf{Notation} & \textbf{Range} \\
\midrule
learning\_rate\_init & \{0.001, 0.003, 0.01, 0.03\} \\
weight\_decay\_init & \{0, 1e-5, 5e-5, 1e-4, 5e-4, 0.001, 0.005\} \\
epoch\_init & \{500, 1000\} \\
patience\_init & \{50, 100, 200, 400\} \\
\midrule
structural\_dim & \{0, $2^6$, $2^7$, ..., $2^{13}$\} \\
hidden\_dim & \{512\} \\
embedding\_dim & \{128\} \\
learning\_rate & \{0.0003, 0.001, 0.003, 0.01, 0.03\} \\
weight\_decay & \{0, 5e-6, 5e-5, 5e-4, 0.001, 0.005, 0.01\} \\
epoch & \{2000\} \\
patience & \{50, 100, 200, 400\} \\
order & \{1, 2\} \\
$\beta$ & \{0.1, 1, 10\} \\
$\tau$ & \{0.1, 0.2, 0.5\} \\
$\lambda$ & [0.8, 1.2] \\
$\kappa$ & [0, 8] \\
$l^{hm}$ & [0, 8] \\
$l^{ht}$ & [0, 8] \\
\bottomrule
\end{tabular}
\end{center}
\end{table}

\subsection{Performance}
\label{sec:perf}
Table \ref{tbl:per} shows the semi-supervised node classification performance results of all the methods on 9 benchmark datasets. 
We highlight the top-rank and rank two results among all methods in bold and underlining respectively for all datasets.
From the table, we have the following observations:

\textbf{MLP} shows the basic baseline performance, which only uses the feature of node ego for classification. 
It performs well on most heterophilous datasets especially on Actor, indicating the important role of ego feature for node classification on both homophilous and heterophilous datasets.

\textbf{General GNN} methods aggregate the neighbors' information to the central node, which brings performance improvement on homophilous datasets.
However, this also leads to the pool performance on heterophilous datasets since a large number of heterophilous neighbors can contaminate the node representation.
As for the improvement on Chameleon and Squirrel compared with MLP, we believe that the neighbors' information contributes much more than the central node to classification in these two datasets.
Thus the neighbors' information, no matter homophilous or heterophilous, can achieve higher performance.
Compared with GCN and GAT, GCNII generally performs better since the initial residual and identity mapping mechanisms implicitly combine intermediate representations and reduce the influence of neighbors, which are fit for heterophilous graphs. 

Predictably, the \textbf{heterophilous GNN} methods perform relatively well on the heterophilous datasets.
As early methods, MixHop, H$_2$GCN, UGCN, GPR-GNN and WRGAT seek for higher homophily while reducing the negative impact of heterophilous neighbors. 
Thus, they also show a slight improvement in homophilous datasets compared with primary HomoGNNs.
However, there are still quite a number of heterophilous neighbors which keep them away from the best performance.
LINKX achieves excellent performance on Squirrel but works badly on others. 
This is probably because the mechanism of separating then mixing adjacency and feature information is more applicable to Squirrel.
Further, GGCN, ACM-GCN and GloGNN utilize the difference between the central node and heterophilous neighbors, which brings additional information for classification.
As a result, they achieve better performance in both heterophilous and homophilous datasets. 

\textbf{\model}achieves the best results in most heterophilous datasets except Actor, demonstrating the effectiveness of semantic-aware neighborhood partition and the heterophilous neighbor distribution modeling. 
For Actor, we believe the unsatisfactory result is due to the inaccurate neighborhood partition which is limited by the classification performance.
For homophilous datasets Cora, Citeseer and Pubmed, the helpful information from heterophilous neighbors is relatively little, which is further shown in Sec \ref{sec:mi}.
The detailed reason is analyzed in Sec \ref{sec:tsne}.
Thus, the performance of \model on homophilous datasets is not the best but also reaches the first echelon.

\subsection{Ablation Study}

\begin{table*}[t]
% \tiny
% \scriptsize
\footnotesize
\renewcommand{\arraystretch}{1.2}
\caption{Ablation study of HDP's main components on representative datasets.}
\label{tbl:ablation}
\begin{center}
\begin{tabular}{c|ccc|c}
\toprule
\textbf{Methods} & \textbf{Wisconsin} & \textbf{Actor} & \textbf{Squirrel} & \textbf{Citeseer}  \\ 
\midrule
Init.            & 86.86 ± 3.29 & 37.24 ± 0.95 & 61.62 ± 1.71 & 75.94 ± 1.21 \\ 
\model w/o Init. & 62.75 ± 13.15 & 26.94 ± 3.14 & 31.60 ± 3.24 & 72.87 ± 2.22\\
\model w/o Homo. & 67.45 ± 5.13 & 35.55 ± 0.82 & 61.05 ± 1.50 & 46.02 ± 5.88 \\
\model w/o Hete. & 86.47 ± 4.25 & 36.79 ± 0.61 & 61.60 ± 1.81 & 77.07 ± 1.51 \\
\model w/o SSE.  & 85.10 ± 2.51 & 37.26 ± 0.67 & 59.75 ± 2.24 & 76.58 ± 1.47 \\
\model w/o SMP.  & 86.86 ± 4.96 & 36.44 ± 0.91 & 55.98 ± 1.75 & 75.00 ± 1.78 \\
\model w/o TPC.  & 74.90 ± 16.28 & 36.36 ± 1.03 & 55.55 ± 12.69 & 73.76 ± 1.53 \\
\model w/o Upd.  & 87.84 ± 2.60 & 37.18 ± 0.73 & 61.97 ± 1.59 & 76.96 ± 1.43\\
\model           & \textbf{88.82 ± 3.40} & \textbf{37.26 ± 0.67} & \textbf{62.07 ± 1.57} & \textbf{77.10 ± 1.56} \\
\bottomrule
\end{tabular}
\end{center}
\end{table*}

\model contains some important components that may have a significant impact on the classification performance.
To show the contribution of each component to the model, we conduct an ablation study on four representative datasets.
Specifically, we explore the role of the assignment initialization module (Init), semantic structural encoding (SSE), homophilous representations (Homo), heterophilous distribution representations (Hete), semantic-aware message passing mechanism (SMP), trusted prototype contrastive loss (TPC) and the update strategy (Upd).
"Init." denotes the results of the assignment initialization module.
For \model without assignment initialization, we use a randomly initialized MLP without training to construct initial assignments.
Meanwhile, for \model without homo-/hetero-philous representations, we only pass the other one as the input of the classifier.
The results are shown in Table \ref{tbl:ablation}.
From the overall level, all seven components have a positive contribution to the model. 
Specifically, we have the following observations and analysis:
\begin{itemize}
    \item The \textbf{assignment initialization module} achieves satisfactory results as a separate module and provides a good foundation for HDP. Without initialization, \model has some obvious performance reduction since the initial partition could be extremely inaccurate, which leads to error accumulation. Fortunately, the update strategy can gradually fix some errors and thus avoid the model collapse.
    \item Relatively speaking, the \textbf{homophilous representations} provide more performance gain than the heterophilous distribution representations, which fits the intuition. Meanwhile, the \textbf{heterophilous neighborhood representations} are also effective on heterophilous datasets especially Squirrel since the absence of homophilous representations only brings slight performance reduction.
    \item The \textbf{semantic structural encoding} provides additional topology and semantic information for representation learning thus improving the performance. Distinctively, it doesn't seem to be helpful to Actor, since the 0-dimension structural embedding performs best. 
    \item The \textbf{semantic-aware message passing} propagates representations along homophilous edges and overcomes the limited neighborhood distribution of a single node caused by the sparse nature of graphs.
    This produces class-unified and more discriminative representations.
    \item The \textbf{trusted prototype contrastive loss} plays a key role in the overall model since it brings discriminability for both homophilous representations and heterophilous distribution representations via the orthogonality-oriented constraint. Without TPC loss, the premise of heterophilous distribution modeling won't be hold, which leads to significant performance degradation.
    \item During the training process, the \textbf{update strategy} creates a virtuous cycle between neighborhood partitions and assignment accuracy. The performance improvement also shows the effectiveness of the mutually enhanced optimization between representation learning and neighborhood partition.
\end{itemize}

\subsection{The Results of Neighborhood Partition}
\begin{figure}[t]
\centerline{\includegraphics[width=0.95\textwidth]{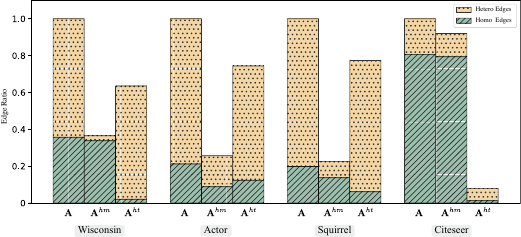}}
\caption{
The visualization of neighborhood partition results on representative datasets. 
The 3 columns of each dataset denote the whole neighborhood, the homophilous and heterophilous neighborhoods partitioned by \model respectively.
}
\label{fig:partition}
\end{figure}
We show the results of neighborhood partition in Figure \ref{fig:partition}.
The first column of each dataset shows the ratio of homophilous edges and heterophilous edges in the whole neighborhood, where the dividing line corresponds to the homophily ratio of the graph.
Note that the original graphs are processed by some operations such as undirected graph conversion and adding self-loop.
Thus the homophily ratio of the processed graph may be different from the original graph.
The second and third columns denote the partitioned homophilous and heterophilous neighborhoods respectively.

Although the accuracy of neighborhood partition is limited by incomplete labels, it still shows great power to handle heterophily thanks to the pseudo assignments.
For Wisconsin, the estimated homophily ratio is almost accurate and the partition result is also quite correct thanks to the high accuracy of assignments. 
Actor and Squirrel have similar heterophily levels, but their partition results are quite different because of the classification accuracy.
For Actor, although the heterophily of the homophilous neighborhood has been reduced, it still suffers from the limitation of unsatisfactory accuracy.
As a result, a large number of homophilous neighbors are partitioned as heterophilous, which leads to inaccurate distribution modeling and further affects the classification performance as we analyzed in Sec \ref{sec:perf}.
For Squirrel, some homophilous edges are incorrectly partitioned, but the homophilous neighborhood is getting better since the heterophily level is reduced.
For Citeseer, there is a gap between the estimated homophily ratio and the truth.
But it is also effective since the homophily ratio becomes higher in the homophilous neighborhood and very low in the heterophilous neighborhood.
To sum up, the adaptive neighborhood partition mechanism can adapt to different levels of heterophily and produce high-quality neighborhoods.

\subsection{Influence of Rescaling Parameter}
\begin{figure}[t]
    \centering
    \subfloat[Squirrel]
    {\includegraphics[width=.47\columnwidth]{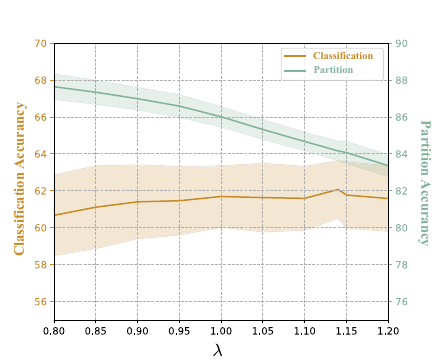}
    \label{fig:lambda_sq}}
    \hspace{10pt}
    \subfloat[Citeseer]
    {\includegraphics[width=.48\columnwidth]{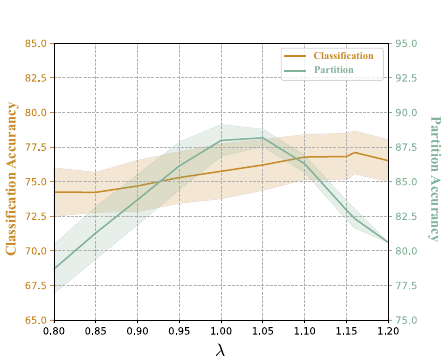}
    \label{fig:labmda_ct}}
    \caption{The variation of partition and classification accuracy while changing rescaling parameter $\lambda$. We calculate the average and standard deviation of 10 split/runs which are shown by the lines and shaded area. }
    \label{fig:lambda}
\end{figure}

The rescaling parameter $\lambda$ controls the trade-off between accuracy and recall of neighborhood partition. 
A low $\lambda$ makes the model choose high-confidence edges as the homophilous neighborhood while abandoning edges that could be homophilous but with relatively low confidence, and vice versa.
To show the impact of $\lambda$, we give the changes of partition accuracy and node classification accuracy concerning $\lambda$ in Fig \ref{fig:lambda}. 
Specifically, partition accuracy is evaluated by regarding the neighborhood partition as a binary classification problem. 

As we expected, the variation of the neighborhood partition shows an inverted U-shaped curve as in Figure \ref{fig:labmda_ct}. 
This also illustrates that the rescaling parameter $\lambda$ is important since estimating graph homophily from the training set may be inaccurate as we said before.
For Squirrel, $\lambda$ needs to be smaller to show the other half of the curve.
As for the classification accuracy, it's quite steady as $\lambda$ changes, which shows the robustness of HDP.
Further, the best point of the two kinds of accuracy is not the same.
A large $\lambda$ seems to be better for the classification accuracy.
This can be attributed to the semantic-aware message passing, for which a relatively complete homophilous neighborhood is desired.

\subsection{Contribution of Homophilous and Heterophilous Representations}
\label{sec:mi}
\begin{figure}[t]
    \centering
    \subfloat[Wisconsin]
    {\includegraphics[width=.35\columnwidth]{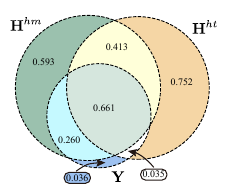}}
    \hspace{10pt}
    \subfloat[Citeseer]
    {\includegraphics[width=.35\columnwidth]{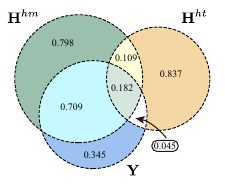}}
    \\
    \subfloat[Actor]
    {\includegraphics[width=.3\columnwidth]{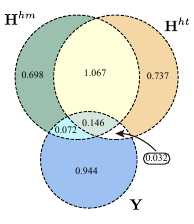}}
    \hspace{20pt}
    \subfloat[Squirrel]
    {\includegraphics[width=.3\columnwidth]{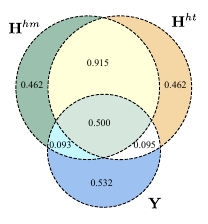}}
    \caption{Mutual information between two kinds of Representations and Labels}
    \label{fig:mi2}
\end{figure}
To intuitively observe the contribution of heterophilous distribution, we estimate the mutual information between two kinds of representations and labels via DIM \cite{Hjelm2018LearningDR}. 
The results are shown in Figure \ref{fig:mi2} as a Venn diagram, where the overlap between two circles denotes the value of corresponding mutual information, and big mutual information value means an important role.
We have the first interesting observation that the overlap between the label circle and others corresponds to the classification performance. The bigger the overlap area is, the bigger the mutual information between labels and representations is, thus the better the classification performance is.
For Wisconsin and Citeseer, homophilous representations play a more important role in node discrimination.
This is consistent with intuition, especially in homophilous datasets like Citeseer.
For Actor and Squirrel, the heterophilous representations show similar contributions with homophilous ones. 
It illustrates that our heterophily modeling is helpful in handling heterophilous graphs.

\subsection{Visualization of Representation Discriminability}
\label{sec:tsne}

\begin{figure}[t]
    \centering
    \subfloat[$\mathbf{H}$ of Squirrel]
    {\includegraphics[width=.3\columnwidth]{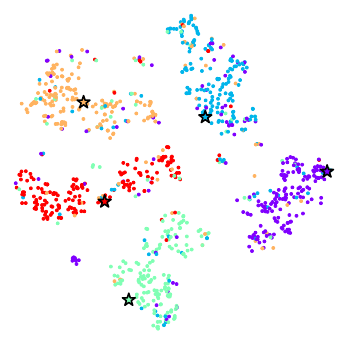}
    \label{fig:sq_h}}
    \hspace{10pt}
    \subfloat[$\mathbf{H}^{hm}$ of Squirrel]
    {\includegraphics[width=.3\columnwidth]{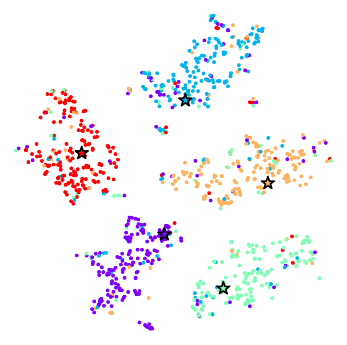}
    \label{fig:sq_hhm}}
    \hspace{10pt}
    \subfloat[$\mathbf{H}^{ht}$ of Squirrel]
    {\includegraphics[width=.3\columnwidth]{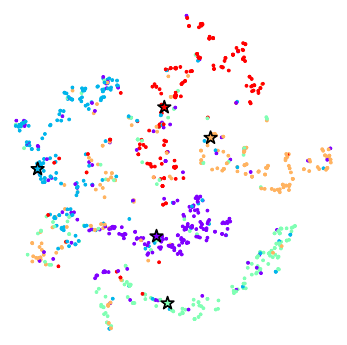}
    \label{fig:sq_hht}}
    \\
    \subfloat[$\mathbf{H}$ of Citeseer]
    {\includegraphics[width=.3\columnwidth]{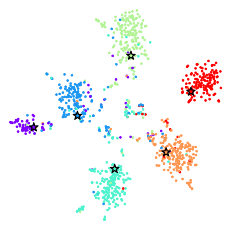}
    \label{fig:ct_h}}
    \hspace{10pt}
    \subfloat[$\mathbf{H}^{hm}$ of Citeseer]
    {\includegraphics[width=.3\columnwidth]{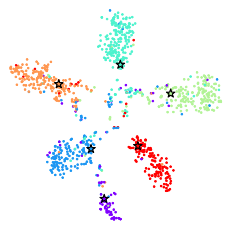}
    \label{fig:ct_hhm}}
    \hspace{10pt}
    \subfloat[$\mathbf{H}^{ht}$ of Citeseer]
    {\includegraphics[width=.3\columnwidth]{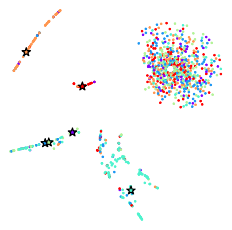}
    \label{fig:ct_hht}}
    \caption{The T-SNE visualization of representations and prototypes.}
    \label{fig:tsne}
\end{figure}
To prove the effectiveness of trusted prototype contrastive loss on the prototype orthogonality, which brings the discriminability to representations, we visualize the node representations and prototypes via T-SNE \cite{van2008visualizing} in Figure \ref{fig:tsne}.
The classes of nodes are shown in different colors while the stars denote the prototypes.
The results of final representations $\mathbf{H}$, homophilous representations $\mathbf{H}^{hm}$ and heterophilous distribution representations $\mathbf{H}^{ht}$ show the effectiveness of HDP, TPC loss and heterophily modeling respectively.

From Figure \ref{fig:sq_h}, \ref{fig:sq_hhm}, \ref{fig:ct_h} and \ref{fig:ct_hhm}, we can see clear boundaries between classes, which indicates the high discriminability of representations. 
Further, they also signify TPC loss well constrains the orthogonality of representations and HDP learns high-quality representations.
In Figure \ref{fig:sq_hht}, the result also shows discrimination.
Note that the heterophilous distribution representations $\mathbf{H}^{ht}$ are constructed without the node's ego feature.
Thus, the discrimination of $\mathbf{H}^{ht}$ illustrates our heterophily modeling is effective and captures additional discriminative information from heterophilous neighbors as we desire.
In Figure \ref{fig:ct_hht}, the result looks like a bit of a mess.
Some representations have clear boundaries between classes while others mix.
This may be due to the quantity of heterophilous neighbors of nodes in the homophilous dataset Citeseer being too small.
Although we solve this problem with the semantic-aware message passing, there still are some classes that don't have clear connection preferences to other classes.
This is a limitation of HDP: if there is no clear connection preference between classes, the heterophily modeling is unable to capture discriminative representations.

\section{Discussion and Conclusion}
In this paper, we study the problem of Heterophlous Graph Neural Networks (HeterGNNs), which is important in real-world graph mining scenarios.
To overcome the shortcomings of existing methods in insufficient neighborhood partition and heterophily modeling, we propose Heterophilous Distribution Propagation for Graph Neural Networks (HDP). 
Specifically, HDP adaptively separates the neighbors into homophilous and heterphilous parts based on the pseudo assignments during training and propagates both homophilous patterns and heterophilous neighborhood distribution with a novel semantic-aware message passing module.
Extensive experiments on 9 real-world datasets demonstrate the effectiveness of the HDP method.
On the other hand, HDP has a limitation that the nodes should have connection preferences to nodes from other classes. Otherwise, the heterophilous distribution will lose its discriminability.
In our future works, we will explore more advanced distribution modeling and more efficient model updating strategies.

\section{Acknowledgements}
This work is supported in part by the National Natural Science Foundation of China (Grant No.62106221, 61972349), Zhejiang Provincial Natural Science Foundation of China (Grant No. LTGG23F030005), and Ningbo Natural Science Foundation (Grant No.2022J183).

\bibliographystyle{elsarticle-num}
\bibliography{reference}

\end{document}